\documentclass[runningheads]{llncs}
\usepackage{graphicx}

\usepackage{tikz}
\usepackage{comment}
\usepackage{amsmath,amssymb} %
\usepackage{color}
\usepackage{graphicx}
\usepackage{comment}
\usepackage{caption}
\usepackage{booktabs}
\usepackage{subfigure}
\usepackage{tabu}
\usepackage{amsmath,amssymb} %
\usepackage{color}
\usepackage{multirow}
\usepackage{threeparttable}
\usepackage[vlined, ruled, linesnumbered]{algorithm2e}
\usepackage{multirow}
\newcommand{\etal}{\textit{et al.}}

\begin{document}
\pagestyle{headings}
\mainmatter
\def\ECCVSubNumber{1094}  %

\title{Efficient Semantic Video Segmentation with Per-frame Inference} %

\titlerunning{Appearing in Proc.\ Eur.\ Conf.\ Comp. Vis. (ECCV), 2020.}
\author{Yifan Liu\inst{1}
\and
Chunhua Shen\inst{1}
\and
Changqian Yu\inst{2,1}
\and
Jingdong Wang\inst{3}
}
\authorrunning{Y. Liu et al.}
\institute{The University of Adelaide, Australia \and
Huazhong University of Science and Technology, China\\ \and
Microsoft Research}

\maketitle

\captionsetup{margin=0.1pt,font=footnotesize,labelfont=bf}

\begin{abstract}

For semantic segmentation, most existing real-time deep models trained with each frame independently may produce inconsistent results
when tested on
a video sequence.  %
A few methods take %
the correlations in the video sequence into account, e.g., by propagating the results to the neighbouring frames using optical flow, or extracting
frame representations
using multi-frame information, which may lead to inaccurate results or unbalanced latency.
In contrast, %
here we explicitly consider the temporal consistency among frames as extra constraints during training and process each frame independently in the inference phase.
Thus no computation overhead is introduced for inference.
Compact models are employed for real-time execution. To narrow the performance gap between compact models and large models, new temporal knowledge distillation methods are designed. Weighing among accuracy, temporal smoothness and %
efficiency, our proposed method outperforms previous keyframe based methods and corresponding baselines which are trained with each frame independently on %
benchmark datasets including %
Cityscapes and Camvid. Code is available at: \url{https://git.io/vidseg}

\end{abstract}

\section{Introduction}
\begin{figure}[thb]
\centering  %
\subfigure[Temporal consistency]{\includegraphics[width=0.52\textwidth]{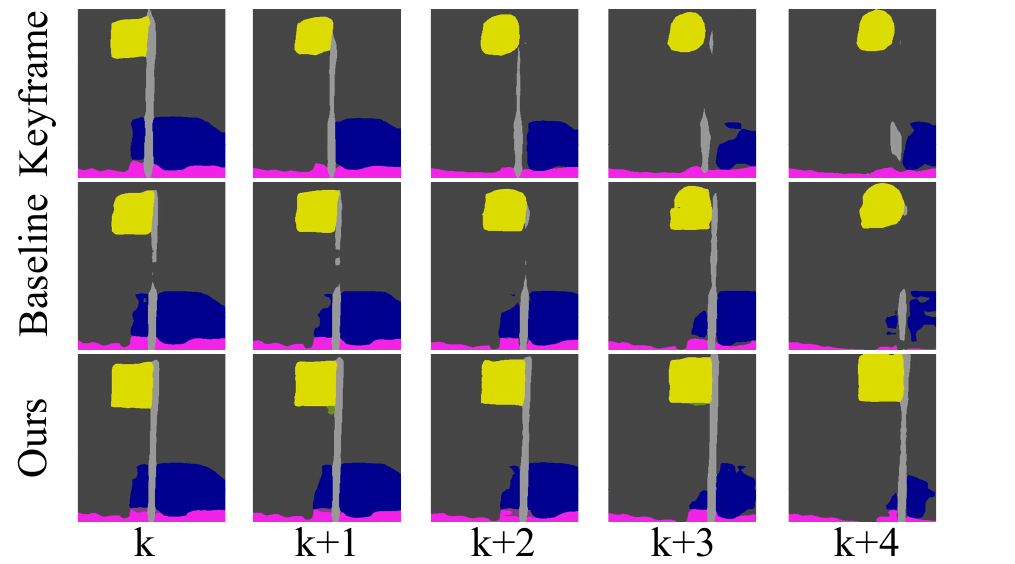}\label{Fig.first}}
\subfigure[Accuracy vs. inference speed.]{\includegraphics[width=0.44\textwidth]{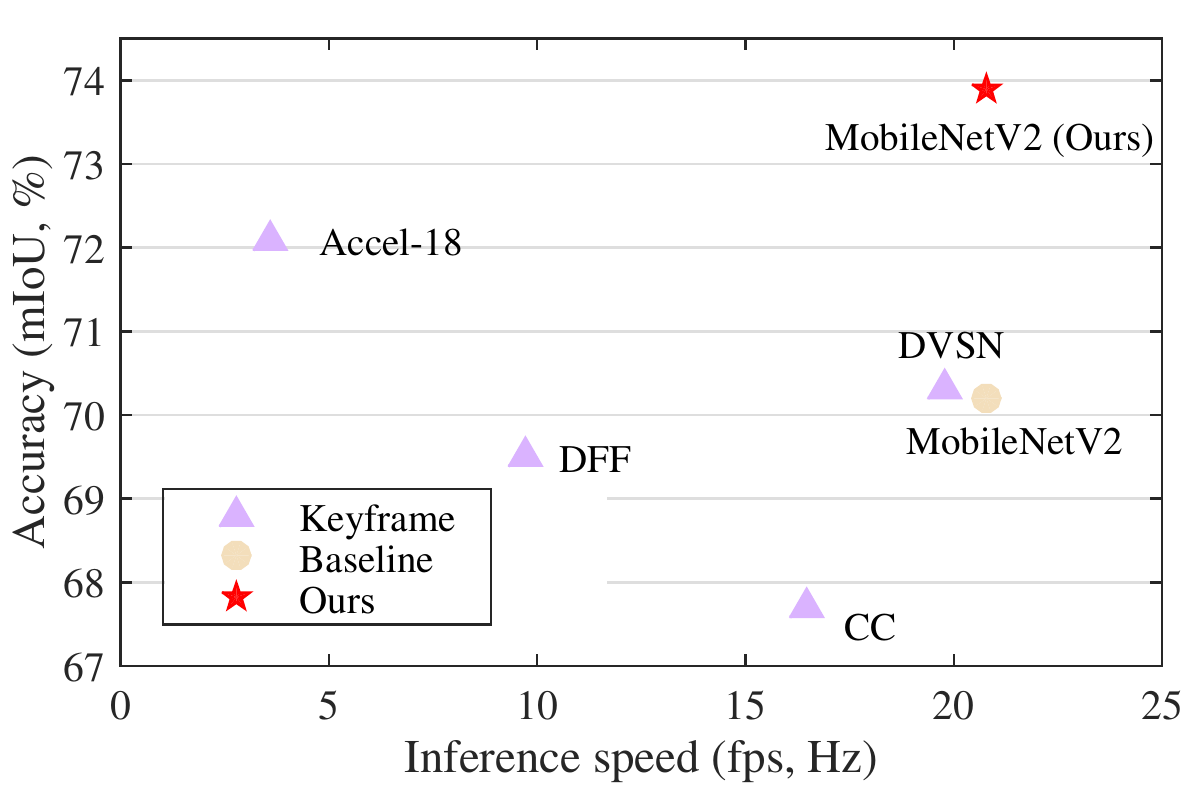}\label{Fig.acc-t}}
\caption{\textbf{(a)} Visualization results on consecutive frames: \textit{Keyframe}: 
Accel18~\cite{jain2019accel} propagates and fuses the results from the keyframe ($k$) to 
non-key
frames ($k+1,\dots$), which %
may lead to
poor results on %
non-key
frames. \textit{Baseline}: PSPNet18~\cite{zhao2017pyramid} trains the model on single frames. Inference %
on single frames separately 
can
produce temporally  inconsistent results. 
\textit{Ours}: training the model with the correlations among frames and inferring %
on single frames separately %
lead to
high quality and smooth results.
\textbf{(b)}
Comparing our enhanced MobileNetV2 model with previous keyframe based methods: Accel~\cite{jain2019accel}, DVSN~\cite{xu2018dynamic}, DFF~\cite{zhu2017deep} and  CC~\cite{shelhamer2016clockwork}. The inference speed is evaluated on a single GTX 1080Ti.}

\end{figure}
Semantic segmentation, a fundamental task in computer vision, aims to assign a semantic label to each pixel in an image. In recent years, the development of deep learning has brought significant success to the task of image semantic segmentation~\cite{zhao2017pyramid,tian2019decoders,chen2018deeplab} on %
benchmark datasets, but often with a high computational cost. This task 
becomes 
computationally 
more expensive when extending to %
video.
For a few real-world applications, e.g., autonomous driving and robotics, it is challenging but crucial to build a fast and accurate video semantic segmentation system.

Previous works for semantic video segmentation can be categorized into two %
groups.
The first group focuses on improving the performance for video segmentation by %
performing
post-processing among frames~\cite{liu2017surveillance}, or employing extra modules to use multi-frames information during inference~\cite{gadde2017semantic}. The high computational cost makes it %
difficult for mobile applications.
The second group %
uses keyframes to avoid processing of each frame, and 
then 
propagate~\cite{zhu2017deep,zhu2018towards,xu2018dynamic} the outputs or the feature maps to other frames (%
non-key
frames)
using 
optical flows. %
Keyframe based methods 
indeed accelerate inference. 
However, it requires different inference time for keyframes and 
non-key
frames, 
leading to an unbalanced latency, thus being not friendly for real-time processing. 
Moreover, accuracy cannot be guaranteed for each frame 
due to the cumulative warping error, for example, the first row in Figure~\ref{Fig.first}.

Efficient semantic segmentation methods on 2D images~\cite{%
mehta2018espnet,yu2018bisenet,orsic2019defense} have draw much attention recently. Clearly, applying compact networks to each frame of %
a  video 
sequence independently may alleviate the latency and enable 
real-time execution. However, directly training the model on each frame independently 
often
produces 
temporally 
inconsistent results on the video %
as shown in the second row of Figure~\ref{Fig.first}. 
To address the above problems, we explicitly consider the temporal consistency among frames as extra constraints during the training process and employ compact networks with per-frame inference %
to ease the problem of latency and achieve %
real-time %
inference.

A motion guided \textit{temporal loss} is employed with the motivation of assigning a consistent label to the same pixel along the time %
axis. 
A motion estimation network is introduced to predict the motion (e.g., optical-flow) of each pixel from the current frame to the next frame based on the input frame-pair. Predicted semantic labels are propagated to the next frame to supervise predictions of the next frame. %
Thus, 
the temporal consistency is encoded into the segmentation network through this constraint.

To narrow the performance gap between compact models and large models,  we design a new \textit{temporal consistency knowledge distillation} 
strategy 
to help the training of compact models. Distillation methods are widely used in image recognition tasks~\cite{liu2019structured,he2019knowledge,Li2017MimickingVE}, and achieve great success. Different from previous distillation methods,  which only consider the spatial correlations, we embed the temporal consistency into distillation items.  We extract the pair-wise frames dependency by calculating the pair-wise similarities for different locations between two frames, and further encode the multi-frames dependency into a latent embedding by using a recurrent unit, ConvLSTM~\cite{ConvLSTM}. The new distillation methods not only improve temporal consistency but also boost segmentation accuracy.  We also include the spatial knowledge distillation methods~\cite{liu2019structured} of single frames in training to further improve the accuracy.

We evaluate the proposed methods on %
semantic video segmentation benchmarks: Cityscapes~\cite{Cordts2016Cityscapes} and Camvid~\cite{brostow2008segmentation}. 
A few
compact backbone networks, i.e., PSPNet18~\cite{zhao2017pyramid}, MobileNetV2~\cite{Sandler2018MobileNetV2IR} and a lightweight HRNet~\cite{SunZJCXLMWLW19}, are included to verify that the proposed methods can empirically improve the segmentation accuracy and the temporal consistency, without any extra computation and post-processing during inference. The proposed methods also show superiority in the trade-off of accuracy and the inference speed.  For example, with the per-frame inference fashion, our enhanced MobileNetV2~\cite{Sandler2018MobileNetV2IR} can achieve higher accuracy with a faster inference speed compared with state-of-the-art keyframe based methods as shown in Figure~\ref{Fig.acc-t}.
We summarize our main contributions as follows.
\begin{itemize}
    \setlength{\itemsep}{0pt}
    \setlength{\parskip}{0pt}
    \setlength{\parsep}{0pt}
    \item We process semantic video segmentation with compact models by per-frame inference, without introducing  post-processing and computation overhead, 
    enabling  real-time inference  without latency.
    \item We explicitly consider the temporal consistency in the training process by using a temporal loss and 
    newly 
    designed temporal consistency knowledge distillation methods.
    \item Empirical experiment results on Cityscapes and Camvid show that with the help of proposed training methods, the compact models outperform previous state-of-the-art semantic video segmentation methods weighing among accuracy, temporal consistency and inference speed. 
\end{itemize}

\subsection{Related Work}
\noindent\textbf{Semantic Video Segmentation.} Semantic video segmentation requires dense labeling for all pixels in each frame of a video sequence into a few semantic categories.
Previous work can be summarized into two streams.

The first one focuses on improving the accuracy by exploiting the temporal relations and the unlabelled data in the video sequence. Nilsson and Sminchiesescu~\cite{nilsson2018semantic} employ a gated recurrent unit to propagate semantic labels to unlabeled frames. Other works like NetWarp~\cite{gadde2017semantic}, STFCN~\cite{fayyaz2016stfcn}, and SVP~\cite{liu2017surveillance} also employ optical-flow or recurrent units to fuse the results of several frames during inferring to improve the segmentation accuracy. Recently, Zhu \etal ~\cite{zhu2019improving} propose to use a motion estimation network to propagate
labels to unlabeled frames as data augmentation and achieve state-of-the-art performance with the segmentation accuracy. These methods can achieve significant performance but
can be difficult to be deployed
on mobile devices.

The second line of works
pay attention to reduce the computational cost by re-using the feature maps in the neighbouring frames. ClockNet~\cite{shelhamer2016clockwork} proposes to copy the feature map to the next frame directly,
thus reducing  the computational cost.
DFF~\cite{zhu2017deep} employs the optical flow to warp the feature map between the keyframe and %
non-key
frames.  Xu \etal ~\cite{xu2018dynamic} further propose to use an adaptive keyframe selection policy while Zhu \etal~\cite{zhu2018towards} find out that propagating partial region in the feature map can get better performance. Li \etal~\cite{li2018low} propose a low-latency video segmentation network by optimizing both the keyframe selection and the adaptive feature propagation. Accel~\cite{jain2019accel} proposes a network fusion policy to use a large model to predict the keyframe and use a compact one in
non-key
frames.
Keyframe based methods require different inference time and may produce different quantity results between keyframes and
other
frames.
In this work, we solve the real-time video segmentation by per-frame inference with
a
compact network and propose a temporal loss and the temporal consistency knowledge distillation to ensure both
good accuracy
and temporal consistency.

\noindent\textbf{Temporal Consistency.}
Applying image processing algorithms to each frame of a video may lead to inconsistent results. The temporal consistency problem has draw much attention in low-level and mid-level applications, such as task-specific methods including colorization~\cite{levin2004colorization}, style transfer~\cite{gupta2017characterizing}, and video depth estimation~\cite{bian2019unsupervised,bian2020unsupervised} and task
agnostic
approaches~\cite{lai2018learning,yao2017occlusion}.
Temporal consistency is also essential in semantic video segmentation.
Miksik \etal~\cite{miksik2013efficient} employ a post-processing method that learns a similarity function between pixels of consecutive frames to propagate predictions across time. Nilsson and Sminchiesescu~\cite{nilsson2018semantic} insert the optical flow estimation network into the forward pass and employ a recurrent unit to make use of neighbouring predictions. Our method is more efficient than
theirs
as we employ  per-frame inference.
The warped previous predictions work as a constraint
\textit{only} during training.

\noindent\textbf{Knowledge Distillation.}
The effectiveness of knowledge distillation has been verified in classification~\cite{hinton2015distilling,romero2014fitnets,Zagoruyko2016PayingMA}. The output of the large teacher net, including the final logits and the intermediate feature maps, are treated as soft targets to supervise the compact student net.
Previous knowledge distillation methods in semantic segmentation~\cite{he2019knowledge,liu2019structured} design distillation
strategies
only for improving the segmentation accuracy.
To our knowledge, to date no distillation methods consider to improve
temporal consistency. In this work, we focus on encoding the motion information into the distillation
terms
to make the segmentation networks more suitable for the semantic video segmentation tasks.

\section{Approach}

\begin{figure}[htbp]
    \centering
    \includegraphics[width=1.0\textwidth]{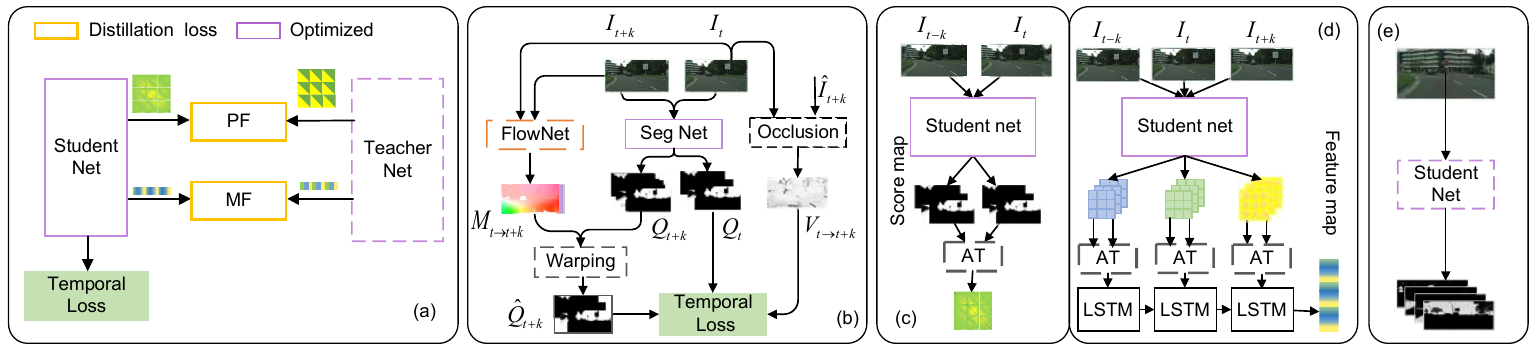}
\     \caption{(a) \textbf{Overall of proposed training scheme}: We consider the temporal information by the temporal consistency knowledge distillation (c and d) and the temporal loss (b) during training. (b) \textbf{Temporal loss (TL)} encode the temporal consistency through motion constraints. Both the teacher net and the student net are enhanced by the temporal loss. (c) \textbf{Pair-wise frame dependency (PF)}: encode the motion relations between two frames. (d) \textbf{multi-frame dependency (MF)}: extract the correlations of the intermediate feature maps among multi-frames.  We only show the forward pass of the student net here and apply the same operations on the teacher net to get the dependency cross frames as soft targets. (e) \textbf{The inference process}. All the proposed methods are only applied during training. We can improve the temporal consistency as well as the segmentation accuracy without any extra parameters or post-processing during inference.}
    \label{fig:overall}
\end{figure}

In this section, we show how we
exploit
the temporal information during training. As shown in  Figure~\ref{fig:overall}(a), we introduce two %
terms:
a
simple
temporal loss (Figure~\ref{fig:overall}(b)) and newly designed temporal consistency knowledge distillation
strategies
(Figure~\ref{fig:overall}(c) and Figure~\ref{fig:overall}(d)). The temporal consistency of the single-frame models can be significantly improved by employing temporal loss. However, if compact models are employed for real-time execution, there is still a performance gap between large models and small ones. We design new
temporal consistency knowledge distillation
to transfer the temporal consistency from large models to small ones. With the help of temporal information, the segmentation accuracy can also be boosted.

\subsection{Motion Guided Temporal Consistency}
\label{sec:TL}

Training semantic segmentation networks independently on each frame of a video sequence often leads to undesired inconsistency. Conventional methods %
include
previous predictions as an extra input, which introduces extra computational cost during inference.
We employ previous predictions as supervised signals to assign consistent labels to each corresponding pixel
along the time axis.

 As shown in Figure~\ref{fig:overall}(b), for two input frames $\mathbf{I}_{t}$, $\mathbf{I}_{t+k}$ from time $t$ and $t+k$, we have:
\begin{equation}
    \ell_{tl}(\mathbf{I}_{t},\mathbf{I}_{t+k})=\frac{1}{N}\sum_{i=1}^{N}V_{t\Rightarrow t+k}^{(i)}\left \|\mathbf{q}_t^{i}-\hat{\mathbf{q}}_{t+k\Rightarrow t}^{i}  \right \|_{2}^{2}
\end{equation}
where $\mathbf{q}_t^{i}$ represents the predicted class probability at the position $i$ of the segmentation map $\mathbf{Q_{t}}$, and $\hat{\mathbf{q}}_{t+k\Rightarrow t}^{i}$ is the warped class probability from frame $t+k$ to frame $t$, by using a motion estimation network(e.g., FlowNet) $f(\cdot)$. Such an $f(\cdot)$ can predict the amount of motion changes in the $x$ and $y$ directions for each pixel: $f(\mathbf{I}_{t+k},\mathbf{I}_{t})=\mathbf{M}_{t \rightarrow t+k}$,
where $\delta i=\mathbf{M}_{t \rightarrow t+k}(i)$, indicating the pixel on the position $i$ of the frame $t$ moves to the position $i+\delta i$ in the frame $t+k$. Therefore,
the segmentation maps between two input frames are aligned by the motion guidance. An occlusion mask $\mathbf{V}_{t\Rightarrow t+k}$ is designed to remove the noise caused by the warping error: $\mathbf{V}_{t\Rightarrow t+k}=\exp(-\left | \mathbf{I}_{t}-
\hat{\mathbf{I}}_{t+k} \right |)$, where $\hat{\mathbf{I}}_{t+k}$ is the warped input frame. We employ a pre-trained optical flow prediction network as the motion estimation net in implementation.
We directly consider the temporal consistency during the training process through the motion guided temporal loss by constraining a moving pixel along the time steps to have a consistent semantic label.
Similar constraints are proposed in image processing tasks~\cite{lai2018learning,yao2017occlusion}, but rarely discussed in semantic segmentation. We find that the straightforward temporal loss can improve the temporal consistency of single-frame models significantly.

\subsection{Temporal Consistency Knowledge Distillation }
\label{sec:tckd}

Inspired by~\cite{liu2019structured}, we build a distillation mechanism to effectively train the compact student net $\mathsf{S}$ by making use of the cumbersome teacher net $\mathsf{T}$. The teacher net $\mathsf{T}$ is already well trained with the cross-entropy loss and the temporal loss to achieve a high temporal consistency as well as the segmentation accuracy. Different from previous single frame distillation methods,
two new distillation
strategies
are designed to transfer the temporal consistency from $\mathsf{T}$ to $\mathsf{S}$: pair-wise-frames dependency (PF) and multi-frame dependency (MF).

\noindent\textbf{Pair-wise-Frames Dependency.}
Following~\cite{liu2019structured}, we denote an attention (AT) operator to calculate the pair-wise similarity map $\mathbf{A}_{\mathbf{X}_1,\mathbf{X}_2}$ of two input tensors $\mathbf{X}_1,\mathbf{X}_2$, where $\mathbf{A}_{\mathbf{X}_1,\mathbf{X}_2} \in \mathbb{R}^{N \times N  \times 1}$ and $\mathbf{X}_1,\mathbf{X}_2 \in \mathbb{R}^{N  \times C} $. For the pixel $a_{ij}$ in $\mathbf{A}$, we calculate the cosine similarity between $\mathbf{x}_{1}^{i}$ and $\mathbf{x}_{2}^{j}$ from $\mathbf{X}_1$ and $\mathbf{X}_2$, respectively: $a_{ij} = {\mathbf{x}_1^i{}^\top \mathbf{x}_{2}^j}/{(\|\mathbf{x}_1^i\|_2\|\mathbf{x}_{2}^j\|_2)}$. It is an efficient and easy way to encode the correlations between two input tensors.

As shown in Figure~\ref{fig:overall}(c), we encode the pair-wise dependency between the prediction of every two neighbouring frame pairs by using the AT operator, and get the similarity map $\mathbf{A}_{\mathbf{Q}_t,\mathbf{Q}_{t+k}}$, where $\mathbf{Q}_t$ is the segmentation map of frame $t$ and $a_{ij}$ of $\mathbf{A}_{\mathbf{Q}_t,\mathbf{Q}_{t+k}}$ denotes the similarity between the class probabilities on the location $i$ of the frame $t$ and the location $j$ of the frame $t+k$. If a pixel on the location $i$ of frame $t$ moves to location $j$ of frame $t+k$, the similarity $a_{ij}$ may be higher. Therefore, the pair-wise dependency can reflect the motion correlation between two frames.

We align the pair-wise-frame (PF) dependency between the teacher net $\mathsf{T}$ and the student net $\mathsf{S}$,
\begin{equation}
    \ell_{PF}(\mathbf{Q_t},\mathbf{Q_{t+k}})=\frac{1}{N^2}\sum_{i=1}^{N}\sum_{j=1}^{N}(a_{ij}^\mathsf{S}-a_{ij}^\mathsf{T})^{2},
\end{equation}
\noindent where $\forall a_{ij}^\mathsf{S} \in \mathbf{A}_{\mathbf{Q}_t,\mathbf{Q}_{t+k}}^\mathsf{S}$ and $
    \forall a_{ij}^\mathsf{T} \in \mathbf{A}_{\mathbf{Q}_t,\mathbf{Q}_{t+k}}^\mathsf{T}$.

\noindent\textbf{Multi-Frame Dependency.} As shown in Figure~\ref{fig:overall}(d), for a video sequence $\mathcal{I}=\{\dots \mathbf{I}_{t-1},\mathbf{I}_{t},\mathbf{I}_{t+1}\dots\}$, the corresponding feature maps
$\mathcal{F}=\{\dots \mathbf{F}_{t-1},\mathbf{F}_{t},\mathbf{F}_{t+1}\dots\}$ are extracted from the output of the last convolutional block before the classification layer. Then, the self-similarity map, $\mathbf{A}_{\mathbf{F}_t,\mathbf{F}_{t}}$, for each frame are calculated by using AT operator in order to: 1) capture the structure information among pixels, and 2) align the different feature channels between the teacher net and student net.

We employ a ConvLSTM unit to encode the sequence of self-similarity maps into an embedding $\mathbf{E}\in \mathbb{R}^{1 \times  D_e}$, where $D_e$ is the length of the embedding space.
For each time step, the ConvLSTM unit takes $\mathbf{A}_{\mathbf{F}_t,\mathbf{F}_{t}}$ and the hidden state which contains the information of previous $t-1$ frames as input and gives an output embedding $\mathbf{E}_t$ along with the hidden state of the current time step. We align the final output embedding~\footnote{The details of calculations in ConvLSTM is referred in~\cite{ConvLSTM}, and we also include the key equations in Section A.2 in Appendix.} at the last time step,   $\mathbf{E}^{\mathsf{T}}$ and  $\mathbf{E}^{\mathsf{S}}$ from $\mathsf{T}$ and $\mathsf{S}$, respectively. The output embedding encodes the relations of the whole input sequence, named multi-frame dependency (MF). The distillation loss based on multi-frame dependency is termed as: $    \ell_{MF}(\mathcal{F})=\left \|\mathbf{E}^{\mathsf{T}}-\mathbf{E}^{\mathsf{S}}\right \|_{2}^{2}$.

The parameters in the ConvLSTM unit are optimized together with the student net. To extract the multi-frame dependency, both the teacher net and the student net share the weight of the ConvLSTM unit. Note that there exists a model collapse point when the weights and bias in the ConvLSTM are all equal to zero. We clip the weights of ConvLSTM between a certain range and enlarges the $\mathbf{E}^{\mathsf{T}}$ as a regularization to prevent the model collapse.

\subsection{Optimization}
\label{sec:svsn}
We pre-train the teacher net with the segmentation loss and the temporal loss to
attain
a segmentation network with a high semantic accuracy and temporal consistency. When optimizing the student net, we fix the weight of the motion estimation net (FlowNet) and the teacher net. These two parts are only used to calculate the temporal loss and the distillation
terms, which can be seen as extra regularization
terms
during the training of the student net. During training, we also employ conventional cross-entropy loss, and the single frame distillation method (SF) proposed in [21] on every single frame to improve the segmentation accuracy. Details can be found in Section A.1. The whole objective function for a sampled video sequence consists of the conventional cross-entropy loss $\ell_{ce}$, the single-frame distillation loss $\ell_{SF}$, temporal loss, and the temporal consistency distillation
terms:
\begin{equation}
\begin{aligned}
        \ell=\sum_{t=1}^{T'}\ell_{ce}^{(t)}+\lambda (\sum_{t=1}^{T}\ell_{SF}^{(t)}+\sum_{i=1}^{T-1}\ell_{tl}(\mathbf{Q}_t,\mathbf{Q}_{t+1})
        +\sum_{i=1}^{T-1}\ell_{PF}(\mathbf{Q}_t,\mathbf{Q}_{t+1})+\ell_{MF}),
\end{aligned}
\end{equation}
\noindent where $T$ is the number of all the frames in one training sequence, and $T'$ is the number of labeled frames. Due to the high labeling cost in semantic video segmentation tasks~\cite{Cordts2016Cityscapes,brostow2008segmentation}, most of the datasets are only annotated with sparse frames. Our methods can be easily applied to the sparse-labeled dataset, because 1) we can make use of large teacher models to generate soft targets; and 2) we care about the temporal consistency between two frames, which can be self-supervised through motion. The loss weight for all regularization
terms
$\lambda$ is set to $0.1$.

After training the compact network, all the motion-estimation net, teacher net, and the distillation modules can be removed. We only keep the student net as the semantic video segmentation network. Thus, both the segmentation accuracy and the temporal consistency can be improved with no extra computational cost in the per-frame inference process.
\section{Implementation details}
\noindent\textbf{Dataset.} We evaluate our proposed method on Camvid~\cite{brostow2008segmentation} and Cityscapes~\cite{Cordts2016Cityscapes}, which are standard benchmarks for semantic video segmentation~\cite{jain2019accel,shelhamer2016clockwork,nilsson2018semantic}. More details of the training and evaluation can be found in Section B of Appendix.
\noindent\textbf{Network structures.}
Different from the keyframe based method, which takes several frames as input during inferring, we apply our training methods to a compact segmentation model with per-frame inference. There are three main parts while training the system:
\begin{itemize}
    \setlength{\itemsep}{0pt}
    \setlength{\parskip}{0pt}
    \setlength{\parsep}{0pt}
    \item A light-weight segmentation network. We conduct most of the experiments on ResNet18 with the architecture of PSPnet~\cite{zhao2017pyramid}, namely PSPNet18. We also employ MobileNetV2~\cite{Sandler2018MobileNetV2IR} and a light-weight HRNet-w18~\cite{SunZJCXLMWLW19} to verify the effectiveness and
generalization ability.
    \item A motion estimation network. We use a pre-trained FlowNetV2~\cite{flownet2-pytorch} to predict the motion between two frames. Because this module can be removed during inferring, we do not need to consider employing a lightweight flownet for acceleration, like in DFF~\cite{zhu2017deep} and GRFP~\cite{nilsson2018semantic}.

    \item A teacher network. We adopt widely-used segmentation architecture PSPNet \cite{zhao2017pyramid} with a ResNet$101$ \cite{He2016DeepRL} as the teacher network, namely PSPNet$101$, which is used to calculate the soft targets in distillation items. We train the teacher net with the temporal loss to enhance the temporal consistency of the teacher.

\end{itemize}

\noindent\textbf{Random sampled policy.}
In order to reduce the computational cost while training video data, and make use of more unlabeled frames, we randomly sample frames in front of the labelled frame, named 'frame\_f' and behind of the labelled frame, named 'frame\_b' to form a training triplet (frame\_f, labelled frame, frame\_b), instead of only using the frames right next to the labelled ones. The random sampled policy can take both long term and short term correlations into consideration and achieve better performance. Training on a longer sequence may
show
better performance with
more expensive computation.

\noindent\textbf{Evaluation metrics.}
We evaluate our method on three aspects: accuracy, temporal consistency, and efficiency. The accuracy is evaluated by %
widely-used
mean Intersection over Union (mIoU) and pixel accuracy for semantic segmentation~\cite{liu2019structured}. We report the model parameters (\#Param) and frames per second (fps) to show the efficiency of employed networks. We follow \cite{lai2018learning} to measure the temporal stability of a video based on the mean flow warping error between every two neighbouring frames.
Different from \cite{lai2018learning}, we use the mIoU score instead of the mean square error to evaluate the semantic segmentation results, and more details can be found
in Appendix.
\begin{table*}[t]
\setlength{\abovecaptionskip}{10pt}
\caption{Accuracy and temporal consistency on Cityscapes validation set. SF: single-frame distillation methods, PF: our proposed pair-wise-frame dependency distillation method. MF: our proposed multi-frame dependency distillation method, TL: the temporal loss. The proposed distillation methods and temporal loss can improve both the temporal consistency and accuracy, and they are complementary to each other.}
\centering
		\setlength{\tabcolsep}{4.6pt}

\begin{tabular}{c|cccc|c|c|c}

\toprule
Scheme index&SF        & PF      & MF        & TL        & mIoU  & Pixel accuracy & Temporal consistency\\ \hline
${a}$&          &           &           &                     & 69.79 & 77.18     & 68.50                \\ \hline
${b}$&\checkmark &           &           &                     & 70.85 & 78.41     & 69.20                \\ \hline
${c}$&          & \checkmark &           &                     & 70.32 & 77.96     & 70.10                \\ \hline
$d$&        &         & \checkmark &                       &    70.38    &        77.99    &    69.78                  \\ \hline
${e}$&          &           &           & \checkmark &         70.67 & 78.46     & 70.46                \\ \hline
${f}$& & \checkmark & \checkmark          &                      & 71.16 &78.69    & 70.21                \\ \hline \hline

${g}$&\checkmark &  &           &\checkmark                      &71.36  &   78.64 &70.13                 \\ \hline
${h}$& & \checkmark &\checkmark           & \checkmark          &71.57           & 78.94 &70.61                 \\ \hline
${i}$&\checkmark & \checkmark & \checkmark &                      & 72.01 & 79.21     & 69.99                \\ \hline
${j}$&\checkmark & \checkmark & \checkmark & \checkmark            & \textbf{73.06} &\textbf{ 80.75}     & \textbf{70.56}                \\ \bottomrule
\end{tabular}

\label{tab:abl}
\end{table*}

\section{Experiments}
\subsection{Ablations}
All the ablation experiments are conducted on the Cityscapes dataset with the PSPNet$18$.

\begin{table}[htbp]
\setlength{\abovecaptionskip}{10pt}
		\setlength{\tabcolsep}{4.2pt}
\caption{Impact of the random sample policy. RS: random sample policy, TC: temporal consistency, TL: temporal loss, Dis: distillation
terms,
ALL: combine TL with Dis. The proposed random sample policy can improve the accuracy and temporal consistency.}
\centering
\begin{tabular}{c|ccc}
\toprule
Method         & RS        & mIoU  & TC   \\ \hline
PSPNet18 + TL  &           & 70.04 & 70.21      \\
PSPNet18 + TL  & \checkmark & 70.67 & 70.46 \\ \hline
PSPNet18 + Dis  &           & 71.24 & 69.48      \\
PSPNet18 + Dis  & \checkmark & 72.01 & 69.99 \\
\hline
PSPNet18 + ALL &           & 72.87 & 70.05 \\
PSPNet18 + ALL & \checkmark & 73.06 & 70.56 \\ \bottomrule
\end{tabular}
\label{tab:RS}
\end{table}
\begin{table}[b]
\setlength{\abovecaptionskip}{10pt}
	\setlength{\tabcolsep}{4.2pt}
\centering
\caption{Influence of the teacher net. TL: temporal loss. TC: temporal consistency. We use the pair-wise-frame distillation to show our design can transfer the temporal consistency from the teacher net.}
\begin{tabular}{l|c|c|c}
\toprule
Method               & Teacher Model        & mIoU  & TC \\ \hline
PSPNet101            &   None                   & 78.84 & 69.71               \\ \hline
PSPNet101 + TL&          None            & 79.53 & 71.68               \\ \hline\hline
PSPNet18             &       None               & 69.79 & 68.50                \\ \hline
PSPNet18             & PSPNet101            & 70.26 & 69.27                \\ \hline
PSPNet18             & PSPNet101 + TL & \textbf{70.32} & \textbf{70.10}                \\
\bottomrule
\end{tabular}
\label{tab:teacher}
\end{table}

\noindent\textbf{Effectiveness of proposed methods.}
In this section, we verify the effectiveness of the proposed training scheme. Both the accuracy and temporal consistency are shown in Table~\ref{tab:abl}. We build the baseline scheme $a$, which is trained on every single labelled frame. Then, we apply three distillation
terms:
the single-frame dependency (SF),  the pair-wise-frame dependency (PF) and multi-frame dependency (MF), separately, to get the scheme ${b}$, ${c}$ and ${d}$. The temporal loss is employed in the scheme ${e}$. Compared with the baseline scheme, all the schemes can improve accuracy as well as temporal consistency.  To compare scheme ${b}$ with ${c}$ and ${d}$, one can see that the newly designed distillation scheme across frames can improve the temporal consistency to a greater extent. From the scheme ${e}$, we can see the temporal loss is most effective for the improvement of temporal consistency. To compare scheme ${f}$ with ${i}$, we can see that single frame distillation methods~\cite{liu2019structured} can improve the segmentation accuracy but may harm the temporal consistency.

To further improve the performance, we combine the distillation terms with the temporal loss and achieve the mIoU of $73.06\%$ and temporal consistency of $70.56\%$. We do not increase any parameters or extra computational cost with per-frame inference. Both the distillation terms and the temporal loss can be seen as regularization terms, which can help the training process. Such regularization terms introduce extra knowledge from the pre-trained teacher net and the motion estimation network. Besides, performance improvement also benefits from temporal information and unlabelled data from the video.

\noindent\textbf{%
Impact of the random sample policy.}
We apply the random sample (RS) policy when training with video sequence in order to make use of more unlabelled images, and capture the long-term dependency. Experiment results are shown in Table~\ref{tab:RS}. By employing the random sampled policy, both the temporal loss and distillation terms can benefit from more sufficient training data in the video sequences, and obtain an improvement on mIoU from $0.24\%$ to $0.69\%$ as well as the temporal consistency from $0.19\%$ to $0.63\%$. We employ such a random sampled policy considering the memory cost during training.
\begin{table}[h]
\setlength{\abovecaptionskip}{6pt}
    \setlength{\tabcolsep}{3.2pt}
\caption{We compare our methods with recent efficient image/video semantic segmentation networks on three aspects: accuracy (mIoU,\%), smoothness (TC, \%) and inference speed (fps, Hz). TL: temporal loss, ALL: all proposed terms, TC: temporal consistency, \#Param: parameters of the networks.}
\centering

\begin{tabular}{l|l|c|c|c|c|c|c|c}
\toprule

\multirow{2}{*}{Method} & \multirow{2}{*}{Backbone} & \multirow{2}{*}{\#Params} & \multicolumn{3}{c|}{Cityscapes} & \multicolumn{3}{c}{Camvid} \\\cline{4-9}
\multicolumn{1}{c|}{}                        &                           &                         & mIoU     & TC       & fps    & mIoU    & TC     & fps   \\\hline
             \multicolumn{9}{c}{Video-based methods: Train and infer on multi frames}      \\\hline

  CC~\cite{shelhamer2016clockwork}                                          & VGG16                   & -                       & 67.7     & 71.2     & 16.5     & -       & -      & -       \\
                   DFF~\cite{zhu2017deep}                                         & ResNet101                   & -                       & 68.7     & 71.4    & 9.7     & 66.0   & 78.0   & 16.1    \\
                                     GRFP~\cite{nilsson2018semantic}                                        &ResNet101                   & -                       & 69.4     & -        & 3.2     & 66.1    & -      & 6.4    \\
                   DVSN~\cite{xu2018dynamic}                                         & ResNet101                     & -                       & 70.3     & -        & 19.8     & -       & -      & -       \\

                                     Accel~\cite{jain2019accel}                                       & ResNet101/18                & -                       & 72.1     & 70.3     & 3.6     &66.7    & 76.2   & 7.1    \\\hline
                                              \multicolumn{9}{c}{Single frame methods: Train and infer on each frame independently} \\
                                     \hline
  PSPNet~\cite{zhao2017pyramid} & ResNet101               & 68.1                    & 78.8     &69.7     &  1.7    & 77.6       & 77.1     & 4.1      \\
  SKD-MV2~\cite{liu2019structured}                                     & MobileNetV2               & 8.3                     & 74.5     &68.2     & 14.4     & -       & -      & -       \\
                   SKD-R18~\cite{liu2019structured}                                   & ResNet18                    & 15.2                    & 72.7     & 67.6     &8.0     & 72.3    & 75.4   & 13.3    \\
                   PSPNet18~\cite{zhao2017pyramid}                                    &  ResNet18                    & 13.2                    & 69.8     & 68.5     & 9.5   & -       & -      & -       \\
                   HRNet-w18~\cite{SunXLW19,SunZJCXLMWLW19}                                   & HRNet                     & 3.9                     & 75.6     & 69.1     & 18.9     & -       & -      & -       \\
                   MobileNetV2~\cite{Sandler2018MobileNetV2IR}                                 & MobileNetV2        & 3.2                     & 70.2     & 68.4     & 20.8     & 74.4    & 76.8   & 27.8    \\        \hline

                           \multicolumn{9}{c}{Ours: Train on multi frames and infer on each frame independently} \\

                   \hline

Teacher Net & ResNet101               & 68.1                    & \textbf{79.5}     &\textbf{71.7}     &  1.7    & \textbf{79.4}       &\textbf{78.6}      & 4.1\\
  PSPNet18+TL                                  & ResNet18              & 13.2                    & 71.1     & 70.0     & 9.5   & -       & -      & -       \\
                   PSPNet18+ALL                                    &  ResNet18                    & 13.2                    & 73.1     & 70.6     & 9.5    & -       & -      & -       \\
                   HRNet-w18+TL                                   & HRNet                     & 3.9                     & 76.4     & 69.6     & 18.9     & -       & -      & -       \\
                   HRNet-w18+ALL                                   & HRNet                     & 3.9                     & 76.6     & 70.1     & 18.9     & -       & -      & -       \\
                   MobileNetV2+TL                                 & MobileNetV2           & 3.2                     & 70.7     & 70.4     & 20.8     & 76.3    & 77.6   & 27.8    \\
                   MobileNetV2+ALL                                 & MobileNetV2      & \textbf{3.2}                     & 73.9     & 69.9     & \textbf{20.8}     & 78.2    & 77.9   & \textbf{27.8 }   \\\bottomrule
\end{tabular}
\label{SOTA}
\end{table}

\noindent\textbf{Impact of the teacher net.}
The temporal loss can improve the temporal consistency of both cumbersome models and compact models. We compare the performance of the student net training with different teacher net (i.e., with and without the proposed temporal loss) to verify that the temporal consistency can be transferred with our designed distillation term. The results are shown in Table~\ref{tab:teacher}. The temporal consistency of the teacher net (PSPNet$101$) can be enhanced by training with temporal loss by $1.97\%$. Meanwhile, the mIoU can also be improved by $0.69\%$.  By using the enhanced teacher net in the distillation framework, the segmentation accuracy is comparable ($70.26$ \textbf{vs.} $70.32$), but the temporal consistency has a significant improvement ($69.27$ \textbf{vs.} $70.10$), indicating that the proposed distillation methods can transfer the temporal consistency from the teacher net.

\noindent \textbf{Discussions.}
We focus on improving the accuracy and temporal consistency for real-time models by making use of temporal correlations. Thus, we do not introduce extra parameters during inference. A series of work~\cite{zhao2017icnet,yu2018bisenet,orsic2019defense} focus on designing network structures for fast segmentation on single images and achieve promising results. They do not contradict to our work. We will verify that our methods can generalize to different network structures, e.g. ResNet18, MobileNetV2 and HRNet in the next session. Besides, large models~\cite{zhao2017pyramid,zhu2019improving} can achieve high segmentation accuracy but have low inference speed. The temporal loss is also effective when applying to large models, e.g., our teacher net.

\subsection{Results on Cityscapes}

\noindent\textbf{Comparison with single-frame based methods.}
Single-frame methods are trained and inferred on each frame independently. Directly apply such methods to video sequences will produce inconsistent results. We apply our training schemes to several efficient single-frame semantic segmentation networks: PSPNet18~\cite{zhao2017pyramid}, MobileNetV2~\cite{Sandler2018MobileNetV2IR} and HRNet-w$18$~\cite{SunZJCXLMWLW19,SunXLW19}.
Metrics of
mIoU, temporal consistency, inference speed, and model parameters are shown in Table~\ref{SOTA}.  As Table~\ref{SOTA} shows, the proposed training scheme
works well with a few
compact backbone networks (e.g., PSPNet18, HRNet-w18 and MobileNetV2). Both temporal consistency and segmentation accuracy can be improved
using
the temporal information among frames.

We also compare our training methods with the single-frame distillation method~\cite{liu2019structured}. According to our observation, GAN based distillation methods proposed in~\cite{liu2019structured}%
can
produce inconsistent results. For example, with the same backbone ResNet18, training with the GAN based distillation methods (SKD-R18)
achieves higher mIoU: $72.7$ vs.\  $69.8$, and a lower temporal consistency: $67.6$ vs.\  $68.5$ compared with the baseline PSPNet18, which is trained with cross-entropy loss on each single frame.
We replace the GAN based distillation term with our temporal consistency distillation terms and the temporal loss,
denoted as %
``PSPNet18+ALL''. Both accuracy and smoothness are improved. Note that we also employ a smaller structure of the PSPNet with half channels than in~\cite{liu2019structured}.

\begin{figure}[t]
    \centering
    \includegraphics[width=0.5\textwidth]{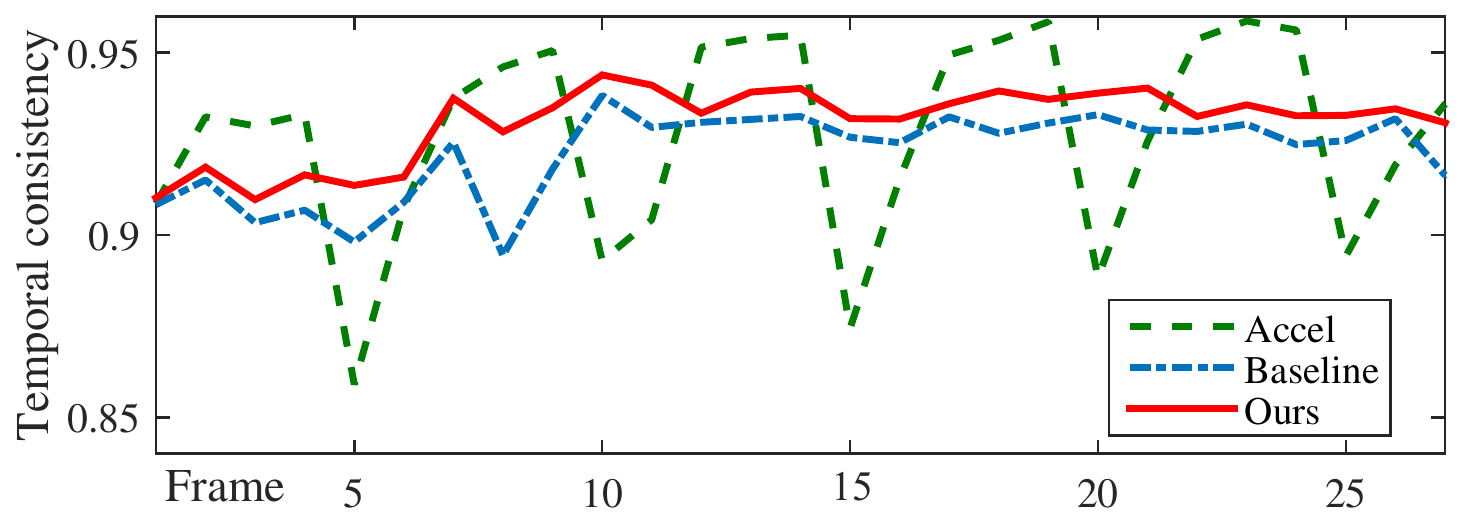}
    \caption{The temporal consistency between neighbouring frames in one sampled sequence on Cityscapes. The keyframe based method Accel
    shows
    severe jitters between keyframes and others.}
    \label{fig:tc_one}

\end{figure}{}

\noindent\textbf{Comparison with video-based methods.}
Video-based methods are trained and inferred on multi frames, we list current methods including keyframe based methods: CC~\cite{shelhamer2016clockwork}, DFF~\cite{zhu2017deep}, DVSN~\cite{xu2018dynamic}, Accel~\cite{jain2019accel}  and multi-frame input method: GRFP~\cite{nilsson2018semantic} in Table~\ref{SOTA}. The compact networks with per-frame inference can be more efficient than video-based methods. Besides, with per-frame inference, semantic segmentation networks have no unbalanced latency and can handle every frame independently.
Table~\ref{SOTA} shows the proposed training schemes can achieve a better trade-off between the accuracy and the inference speed compared with other state-of-the-art semantic video segmentation methods, especially the MobileNetV2 with the fps of $20.8$ and mIoU of $73.9$. Although keyframe methods can achieve a high average temporal consistency score, the predictions beyond the keyframe are in low quality. Thus, the temporal consistency will be quiet low between keyframe and non-key frames, as shown in Figure~\ref{fig:tc_one}. The high average temporal consistency score is mainly from the low-quality predictions on non-key frames. In contrast, our method can produce stable segmentation results on each frame.

\begin{figure*}[tbp]
    \centering
    \includegraphics[width=1.0\textwidth]{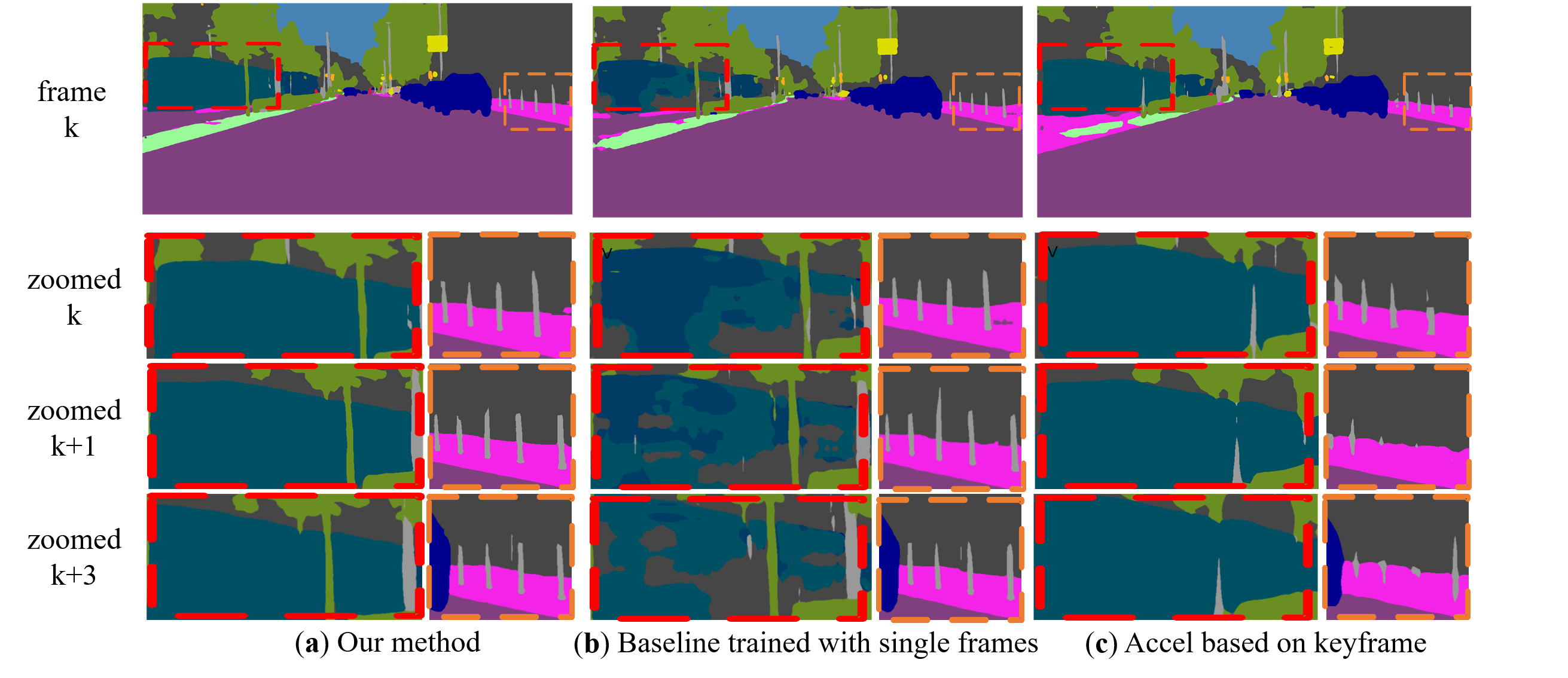}
    \caption{\textbf{Qualitative outputs.}  (\textbf{a}): PSPNet18, training on multi frames and inferring on each frame. (\textbf{b}): PSPNet18, training and inferring on each frame. (\textbf{c}): Accel-18~\cite{jain2019accel}, training and inferring on multiple frames. The keyframe is selected in every five frames. For better visualization, we zoom the region in the red and orange box. The proposed method can give more consistent labels to the moving train and the trees in the red box. In the orange boxes, we can see our methods have similar quantity results in each frame while the keyframe based methods may generate worse results in the frame (e.g., $k+3$) which is far from the keyframe (i.e., $k$).}
    \label{fig:city_vis}
\end{figure*}
\noindent\textbf{Qualitative visualization.}
Qualitative visualization results are shown in Figure~\ref{fig:city_vis}, in which, we can see, the keyframe-based method Accel-18 will produce unbalanced quality segmentation results between the keyframe (e.g., the orange box of $k$) and non-key frames (e.g., the orange box of $k+1$ and $k+3$ ), due to the different forward-networks it chooses. By contrast, ours can produce stable results on the video sequence because we use the same enhanced network on all frames. Compared with the baseline method trained on single frames, we can see our proposed method can produce more smooth results, e.g., the region in red boxes. The improvement of temporal consistency is more clearly shown in the video comparison results.  Moreover, we show a case of the temporal consistency between neighbouring frames in a sampled frame sequence in Figure~\ref{fig:tc_one}. Temporal consistency between two frames is evaluated by the warping pixel accuracy. The higher, the better. The keyframe based method will produce jitters between keyframe and non-key frames, while our training methods can improve the temporal consistency for every frame. The temporal consistency between non-key frames are higher than our methods, but the segmentation performance is lower than ours.

\subsection{CamVid}
We provide additional experiments on CamVid.
We use %
MobileNetV2 as the backbone in the PSPNet.  In Table~\ref{SOTA}, the segmentation accuracy, and the temporal consistency are improved compared with the baseline method. We also outperform current state-of-the-art semantic video segmentation methods with a better trade-off between the accuracy and the inference speed. We use the pre-trained weight from cityscapes following VideoGCRF~\cite{chandra2018deep}, and achieve better segmentation results of $78.2$ vs. $75.2$. VideoGCRF~\cite{chandra2018deep} can achieve $22$ fps with $321\times321$ resolution on a GTX $1080$ card.
We can achieve $78$ fps with the same resolution.
The consistent improvements on both datasets
verify the value of
our training schemes for real-time semantic video segmentation.

\section{Conclusions}
In this work, we have developed real-time video segmentation methods that consider not only accuracy but also temporal consistency. To this end,
we have proposed to use compact networks
with per-frame inference. %
We explicitly consider the temporal correlation during training by using: the temporal loss and the new temporal consistency knowledge distillation.
For inference,
the model
processes
each frame separately, which
does not introduce
latency and avoids post-processing.
The compact networks achieve considerably better temporal consistency and semantic accuracy, %
without introducing
extra computational cost during inference. Our experiments have verified the effectiveness of each component that we have designed. They can improve the performance individually and are  complement to each other.

\textbf{Acknowledgements}
Correspondence should be addressed to CS.
CS was in part supported by ARC DP `Deep learning that scales'.

\appendix

\section*{Appendix}

\section{Details of distillation mechanism}
\subsection{Single frame distillation}
Following Liu et.al.~\cite{liu2019structured}, we employ pixel-wise distillation and pair-wise distillation for each single frame.
For the pixel-wise distillation, we use the class probabilities $Q$ produced from the cumbersome model as {soft targets} for training the compact network.

The loss function based on the Kullback$\textrm{-}$Leibler divergence is given as follows,
\begin{align}
\ell_{pi} = \frac{1}{N} \sum_{i=1}^{N} \operatorname{KL}(\mathbf{q}^s_i \| \mathbf{q}^t_i),
\end{align}
where $\mathbf{q}_i$ represent the class probabilities
of the $i$th pixel of the segmentation map and $N$ is the number of the pixels.

The pair-wise distillation is built on the self-similarity map $\mathbf{A}$ as described in multi-frame dependency. We adopt the squared difference to formulate the pair-wise similarity distillation loss,
\begin{align}
\ell_{pa} = \frac{1}{N^2}
\sum_{i=1}^{N} \sum_{i=j}^{N}
(a^s_{ij} - a^t_{ij})^2.
\end{align}
The similarity between two pixels is simply computed
from the features $\mathbf{f}_i$
and $\mathbf{f}_j$
as $a_{ij} = {\mathbf{f}_i^\top \mathbf{f}_j}/{(\|\mathbf{f}_i\|_2\|\mathbf{f}_j\|_2)}.$
The final loss for sing frame distillation is $\ell_{SF}^t=\ell_{pi}+\ell_{pa}$
\subsection{Multi-frame distillation}
We employ a ConvLSTM~\cite{ConvLSTM} unit to capture the correlations among all frames in a video sequence. The input sequence is consists of the self-similarity maps of the feature map for each frame, $\mathcal{A}=\{\dots \mathbf{A}_{\mathbf{F}_{t-1},\mathbf{F}_{t-1}},\mathbf{A}_{\mathbf{F}_t,\mathbf{F}_t},\mathbf{A}_{\mathbf{F}_t+1,\mathbf{F}_t+1}\dots\}$. For each time step,the key equations are shown in below:
\begin{equation}
\begin{aligned}
\mathbf{i}_t =& \sigma(\mathbf{W}_{ai}\ast \mathbf{A}_{\mathbf{F}_t,\mathbf{F}_t} + \mathbf{W}_{hi}\ast \mathbf{H}_{t-1} + \mathbf{W}_{ei}\circ \mathbf{E}_{t-1} + \mathbf{b}_i) \\
\mathbf{f}_t =& \sigma(\mathbf{W}_{af}\ast\mathbf{A}_{\mathbf{F}_t,\mathbf{F}_t} + \mathbf{W}_{hf}\ast \mathbf{H}_{t-1} + \mathbf{W}_{ef}\circ \mathbf{E}_{t-1}+\mathbf{b}_f) \\
\mathbf{E}_t = &\mathbf{f}_t \circ \mathbf{E}_{t-1} + \\
&\mathbf{i}_t \circ \tanh(\mathbf{W}_{ae} \ast \mathbf{A}_{\mathbf{F}_t,\mathbf{F}_t} + \mathbf{W}_{he} \ast \mathbf{H}_{t-1}+\mathbf{b}_e) \\
\mathbf{o}_t =& \sigma(\mathbf{W}_{ao}\ast \mathbf{A}_{\mathbf{F}_t,\mathbf{F}_t}  +\mathbf{W}_{ho}\ast \mathbf{H}_{t-1} + \mathbf{W}_{eo}\circ \mathbf{E}_{t}  +\mathbf{b}_o) \\
\mathbf{H}_t = &\mathbf{o}_t \circ \tanh({\mathbf{E}_t})
\end{aligned}
\label{eq:convlstm}
\end{equation}

where `$\circ$' denotes the Hadamard product,`$\ast$' denotes the convolution operator, `$\sigma$' is the sigmoid activation function and the activation of input gate $i_t$ controls whether the new input of this time step will be engaged in the memory cell. $\mathbf{f}_t$ controls how much to keep from the past cell status $\mathbf{E}_{t-1}$. $o_t$ decides the propagation from $\mathbf{E}_{t}$ to the hidden state $\mathbf{H}_{t}$. $\mathbf{W}$  and $\mathbf{b}$ represent the trainable parameters in the ConvLSTM unit. We employ the memory state of the final time step $\mathbf{E}_T$ as the distillation item, which contains multi-frame dependency. We align the multi-frame dependency from the teacher net and the student net to enhance the performance of the student net. According to~\cite{ConvLSTM}, the state of ConvLSTM unit can be viewed as the hidden representations of moving objects, therefore the multi-frame dependency distillation can help to transfer the temporal consistency from teacher net to the student net.
\section{Training and evaluation details }
\subsection{Dataset}
Cityscapes~\cite{Cordts2016Cityscapes}
is collected for urban scene understanding
and contains $30$-frame snippets of the street scene with $17$ frames per second.
The dataset contains $5,000$ high quality pixel-level finely annotated images at $20^{th}$ frame in each snippets, which are divided into $2,975$, $500$,  $1,525$ images for training, validation and testing. The CamVid dataset~\cite{brostow2008segmentation} is an automotive dataset. It contains five different videos, which has ground truth labels every $30$ frames.
Three train videos contain $367$ frames, while two test videos contain $233$ frames.
\subsection{Training and inference.}
On Cityscapes, the segmentation networks in this paper are trained by mini-batch stochastic gradient descent (SGD) for $200$ epochs. We sample $8$ training triplets for each mini-batch. The learning rate is initialized as $0.01$ and is multiplied by $(1-\frac{iter}{ max-iter})^{0.9}$.  We randomly cut the images into $769 \times 769$ as the training input. Normal data augmentation methods are applied during training, such as random scaling (from $0.5$ to $2.1$) and random flipping. On Camvid, we use a crop size of $640\times 640$. We use the official implementation of PSPNet in Pytorch\cite{semseg2019} and train all the network with $4$ cards of Tesla Volta $100$.
\subsection{Details of the evaluating temporal consistency}
We follow \cite{lai2018learning} to measure the temporal stability of a video based on the flow warping error between two frames.
Different from \cite{lai2018learning}, we use the mIoU score instead of the mean square error to evaluate the semantic segmentation results
\begin{equation}
    E_{warp}({\mathbf{Q}_{t-1},\mathbf{Q}_{t}})=\frac{\mathbf{Q}_{t}\cap \hat{\mathbf{Q}}_{t-1}}{\mathbf{Q}_{t}\cup \hat{\mathbf{Q}}_{t-1}}
\end{equation}
where $\mathbf{Q}_{t}$ represents for the predict segmentation map of frame $t$ and $\hat{\mathbf{Q}}_{t-1}$ represents for the warped segmentation map from frame $t-1$ to frame $t$.
We calculate a statistical average warp IoU on each sequence, and using an average mean on the validation set to evaluate the temporal stability:
\begin{equation}
    E_{warp}=\frac{1}{N}\sum_{i=1}^{N}\frac{\mathcal{Q}^{i}\cap \hat{\mathcal{Q}}^{i}}{\mathcal{Q}^{i}\cup \hat{\mathcal{Q}}^{i}}
\end{equation}
where $\mathcal{Q}=\{\mathbf{Q}_2,\dots,\mathbf{Q}_{T}\}$ and $\hat{\mathcal{Q}}=\{\hat{\mathbf{Q}}_1,\dots,\hat{\mathbf{Q}}_{T-1}\}$. $T$ is the total frames of the sequence and $N$ is the number of the sequence.
On Cityscapes~\cite{Cordts2016Cityscapes}, we random sample $100$ video sequence from the validation set, which contains $3000$ images to evaluate the temporal stability. On Camvid~\cite{brostow2008segmentation}, we evaluate the temporal stability of the video sequence `seq05' from the test set.

\section{Description of videos and visualization results}
\begin{figure}[htp]
\setlength{\belowcaptionskip}{-0.2cm}
\centering
\subfigure[An example in val.mp4]{
\includegraphics[width=0.4\textwidth]{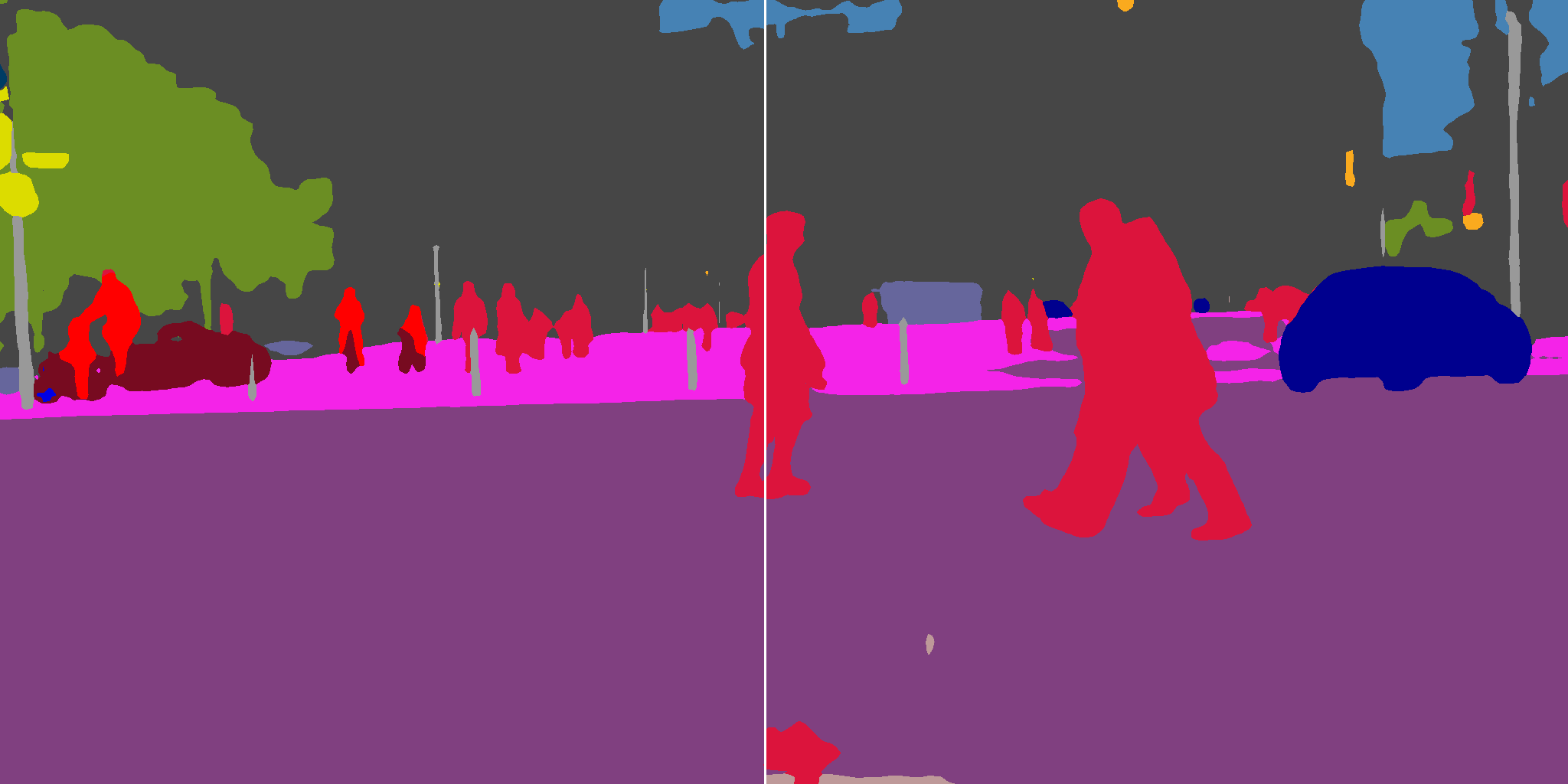}}
\subfigure[An example in demo\_seq00.mp4]{
\includegraphics[width=0.4\textwidth]{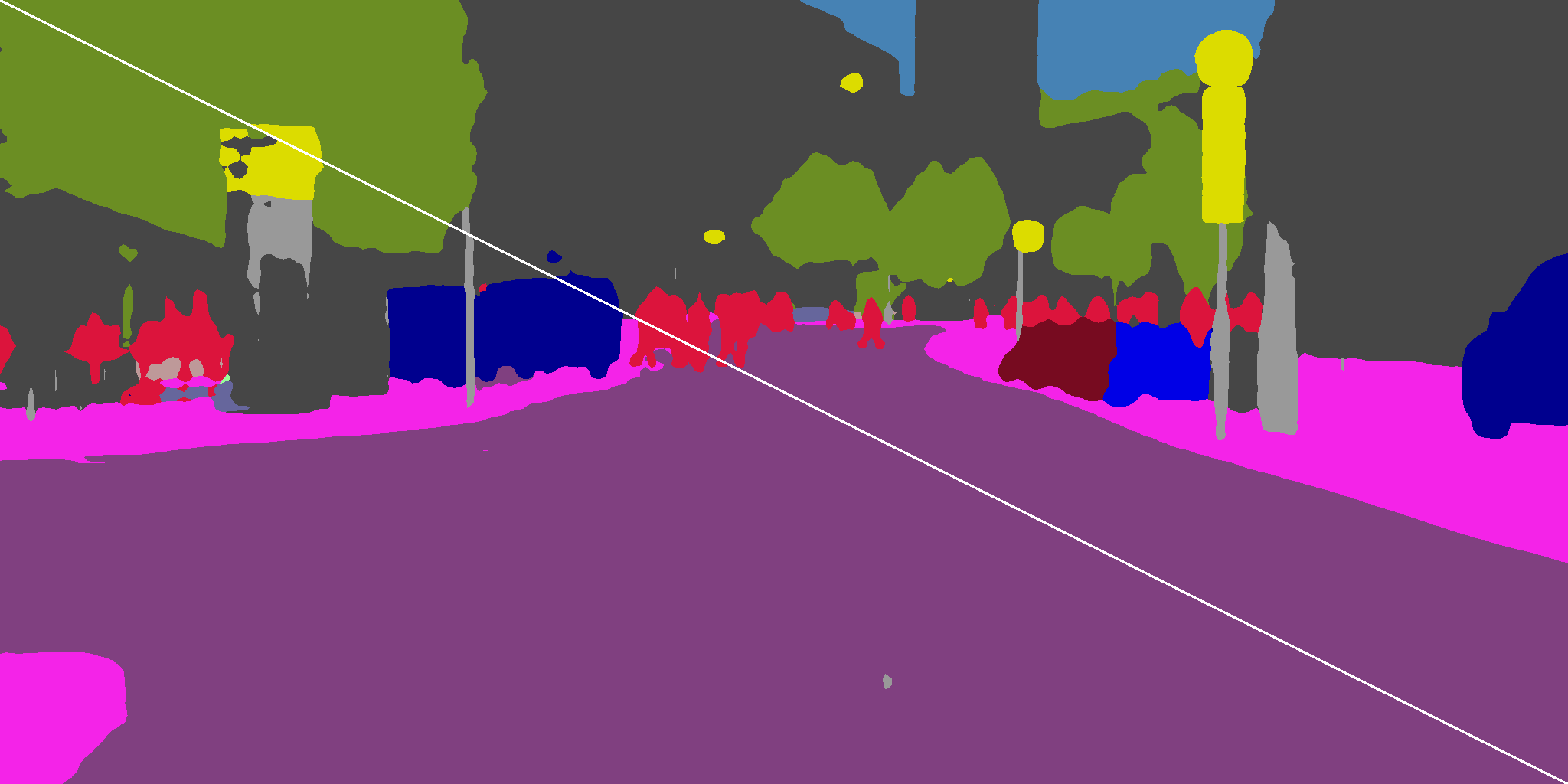}}
\caption{We use a white line to divide the whole scene into two parts. `val.mp4' is the sampled from the validation set we use to evaluate the temporal consistency. In `val.mp4', our method is shown on the left of the line while the baseline method is on the right. `demo\_seq00.mp4' is the prediction results on the provided demo video `sequence00' in the Cityscapes dataset, and our method is above the line while the baseline is below the line.}

\label{fig:demo examples}
\end{figure}

\begin{figure*}[htp]
    \centering
    \includegraphics[width=0.9\textwidth]{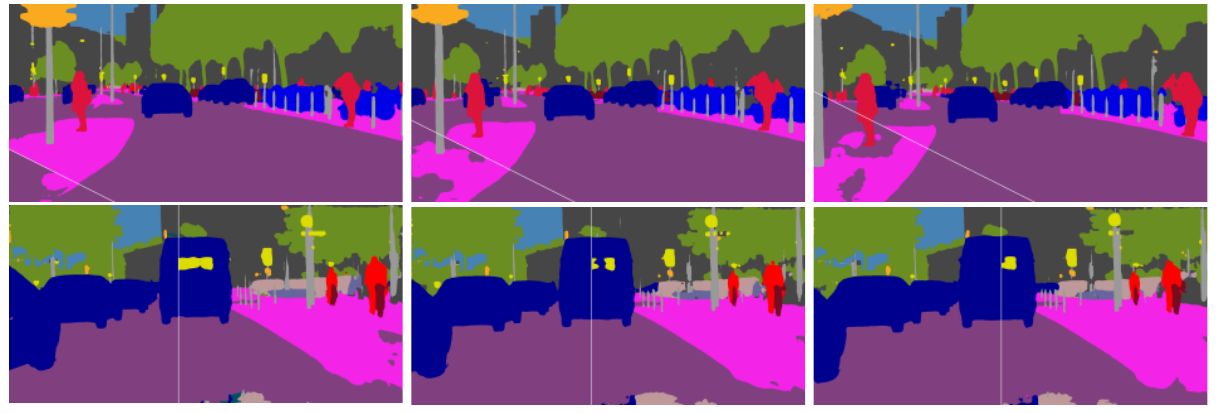}
    \caption{Consecutive frames in two videos. \textbf{First row}:`demo\_seq00.mp4'. Our results are on the top right. \textbf{Second row}:`val.mp4'. Our results are on the left.  More results can be found in the supplementary videos.}
    \label{fig:samples}
\end{figure*}

\begin{figure*}[htp]
    \centering
   \subfigure[frame k]{\includegraphics[width=0.8\textwidth]{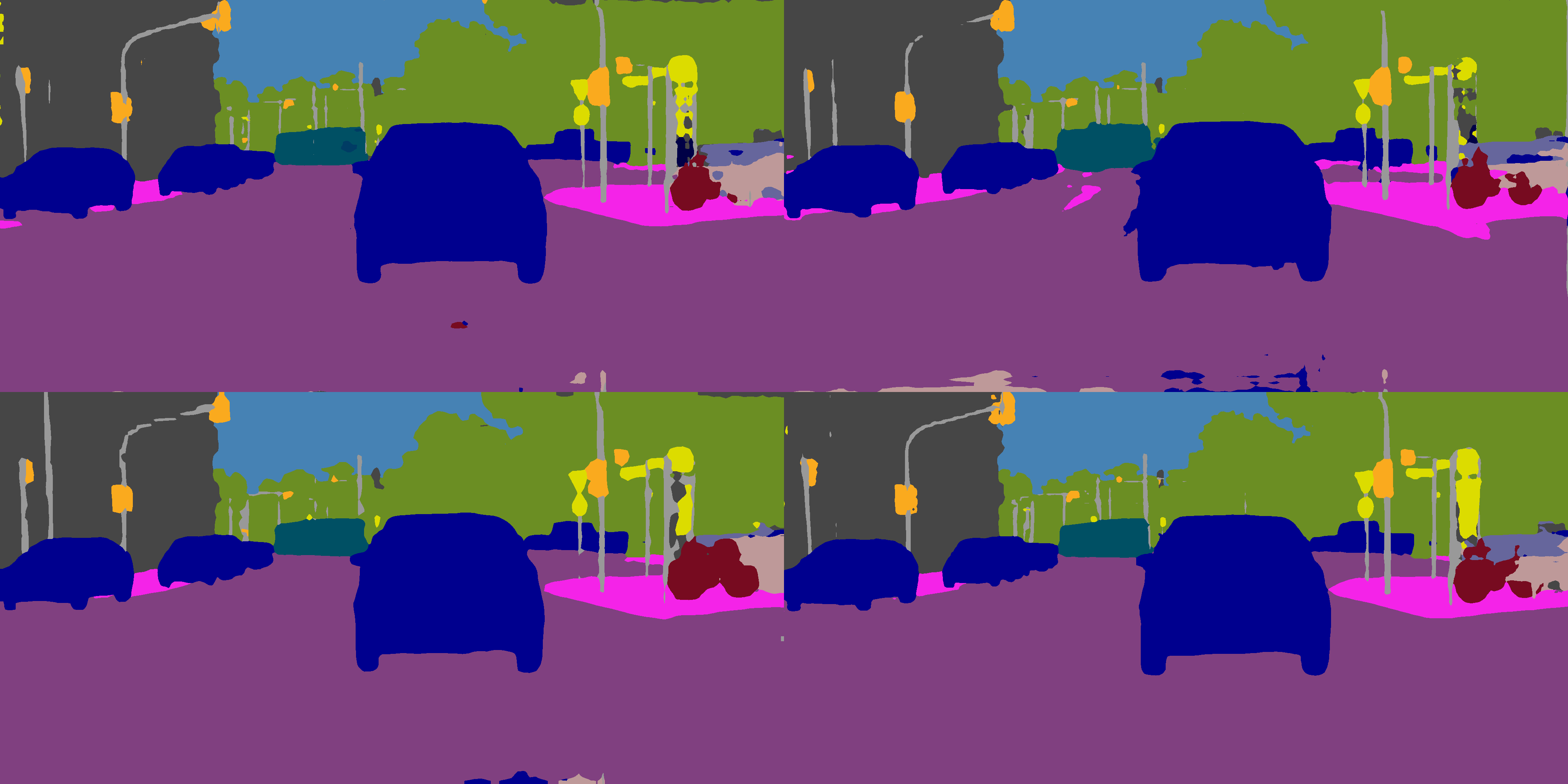}}
   \subfigure[frame k+1]{\includegraphics[width=0.8\textwidth]{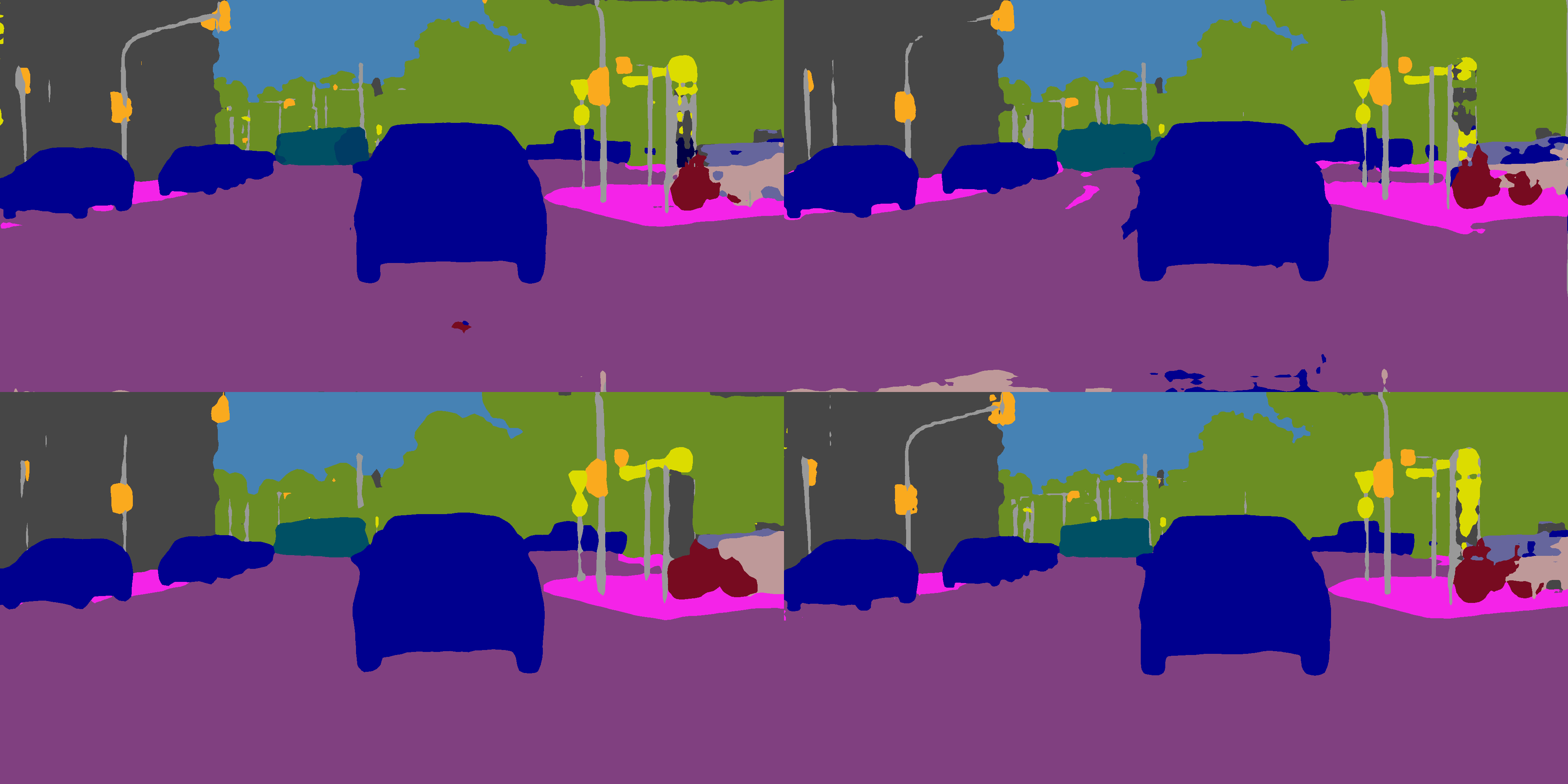}}
   \subfigure[frame k+2]{\includegraphics[width=0.8\textwidth]{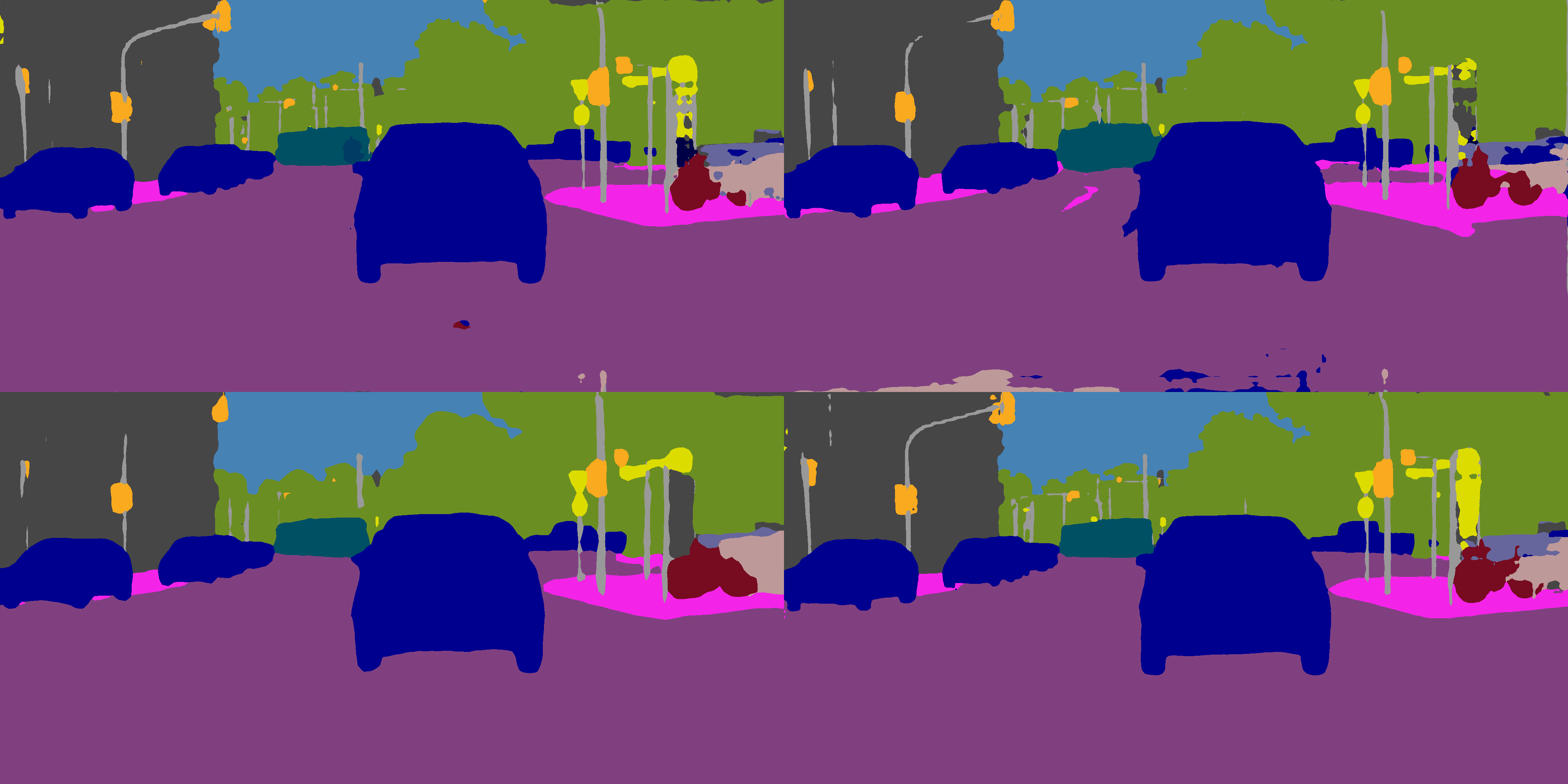}}
    \caption{Consecutive frames in 'Baseline\_SKD\_Accel\_Ours.mp4'. \textbf{Top left}: Baseline.\textbf{Top right}: SKD~\cite{liu2019structured}. \textbf{Bottom left}: Accel~\cite{jain2019accel}. \textbf{Bottom right}: Ours. There are jitters between keyframe and normal frame in the results sequence of Accel.
    More results can be found in the supplementary videos.}
    \label{fig:four}
\end{figure*}

We include three videos in the supplementary materials, named `demo\_seq00.mp4', `val.mp4', and 'Baseline\_SKD\_Accel\_Ours.mp4' to show the improvement of the temporal consistency. Sampled frames are shown in Figure ~\ref{fig:demo examples}, Figure~\ref{fig:samples} and Figure~\ref{fig:four}. From the video, we can see that the proposed method can improve the accuracy and the temporal consistency compared with the baseline models. We can also observe that in some situations, both our method and the baseline method will produce inconsistent predictions. From the comparison with Accel~\cite{jain2019accel} and SKD~\cite{liu2019structured}, we can see that keyframe based methods suffer from the jitters while GAN based distillation methods will produce inconsistent results. Our method can produce stable and smooth sequence with high accuracy.

\begin{figure*}
    \centering
    \includegraphics[width=0.9\textwidth]{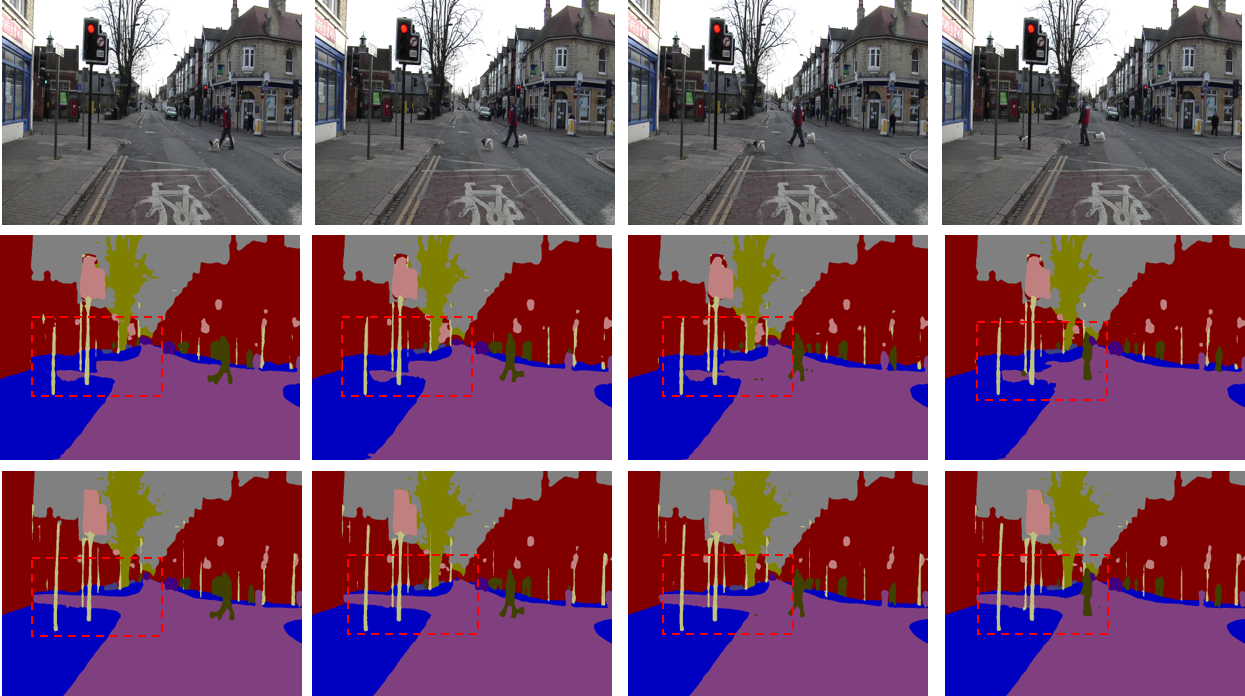}
    \caption{Consecutive frames in Camvid dataset. \textbf{First row}: input frames. \textbf{Second row}: MobileNet trained with cross-entropy loss. \textbf{Third row}:MobileNet trained with the temporal loss and  distillation items. In the baseline method, the region in the red box keep changing while the proposed method can produce similar results on the still stuff.}
    \label{fig:samples_camvid}
\end{figure*}

Figure~\ref{fig:samples_camvid} shows some segmentation results on CamVid dataset.
We can observe that the proposed method outperforms the baseline method in the red region.

{}

{}

{}

\section{Results on each class}
We compare our method with the baseline methods of PSPNet18 in terms of the accuracy and temporal consistency of each class on Cityscapes. The results are shown in Table~\ref{fig:classes}. For the moving objects with regular structures, e.g. `train', `bus', both segmentation accuracy and temporal consistency are improved significantly. For the `road', `sidewalk' and `terrain', the temporal consistency are also improved although the accuracy only have limited improvements.
\begin{table}[htb]
\scriptsize
\vspace{-2em}
\caption{Accuracy (mIoU, \%) and temporal consistency  (TC, \%) for each class on Cityscapes. Baseline: PSPNet18 trained on each frame independently. Ours: PSPNet18 trained with temporal loss and distillation items.}
\begin{tabular}{l|l|c|c|c|c|c|c|c|c|c|c}
\toprule
\multicolumn{2}{c|}{Class Name}                                           & \multicolumn{1}{c|}{road}   & \multicolumn{1}{c|}{sidewalk} & building & wall   & fence  & pole   & tra. light & tra. sign & vegetation & terrain  \\\hline
\multicolumn{1}{c|}{\multirow{2}{*}{mIoU}} & \multicolumn{1}{c|}{Baseline} & \multicolumn{1}{c|}{97.0} & \multicolumn{1}{c|}{78.7}   & 90.1   & 41.8 & 54.7 & 50.3 & 63.6        & 72.0       & 90.8     & 60.0   \\
\multicolumn{1}{c|}{}                      & Ours                         & \textbf{97.2}                     & \textbf{79.4}                       & \textbf{91.0}   & \textbf{49.8} & \textbf{57.4} &\textbf{ 53.1} & \textbf{67.0}        & \textbf{73.6 }      & \textbf{91.0}     & 60.0  \\\hline
\multicolumn{1}{c|}{\multirow{2}{*}{TC}}                       & Baseline                     & 97.2                     & 80.2                       & 91.2   & \textbf{50.0} & 62.1 & 42.6 & 47.2        & 52.6       & 91.7     & 72.0  \\
                                          & Ours                         &\textbf{ 97.7}                     & \textbf{81.4}                       & \textbf{91.6}   & 49.6 & \textbf{62.6} & \textbf{43.9} & \textbf{48.5 }       & \textbf{53.2}       & \textbf{91.9}     & \textbf{73.3}  \\
                                          \hline

\multicolumn{2}{l|}{Class Name}                                           & sky                        & person                       & rider    & car    & truck  & bus   & train         & motorbike  & bicycle    & mean     \\\hline
\multicolumn{1}{c|}{\multirow{2}{*}{mIoU}}                                   & Baseline                     & 92.8                     & 75.8                       & 52.7   & 91.6 & 61.4 & 77.1 & 56.9        & 46.9       & 71.8     & 69.8  \\
                                          & Ours                         & \textbf{93.1}                     & \textbf{77.1}                       &\textbf{ 57.1}   & \textbf{92.1} & \textbf{65.5} &\textbf{ 82.2} & \textbf{73.1}        & \textbf{55.6}       & \textbf{72.8}     & \textbf{73.1}   \\\hline
\multicolumn{1}{c|}{\multirow{2}{*}{TC}}                      & Baseline                     & 92.8                     & 68.7                       & 28.7   & 86.4 & 74.8 & 78.5 & 55.5        & 55.9       & 73.7     & 68.5  \\
                                          & Ours                         & \textbf{93.0}                     &\textbf{ 69.6}                       & \textbf{30.1}   & \textbf{87.0} & \textbf{76.3} & \textbf{82.2} & \textbf{76.4 }      &\textbf{ 57.5}       & \textbf{74.9}     & \textbf{70.6}\\
                                          \bottomrule
\end{tabular}
\label{fig:classes}
\end{table}

\clearpage
\bibliographystyle{splncs04}
\bibliography{1094}

\end{document}